# AUTOPILOT: An Advanced Perception, Localization and Path Planning Techniques for Autonomous Vehicles using YOLOv7 and MiDaS

Mr. Harshkumar Devmurari
*Dept. of CSE(DS)*
*Vidyavardhini's College of Engineering and Technology*
Mumbai, India
harshdevmurari007@gmail.com

Mr. Gautham Kuckian
*Dept. of CSE(DS)*
*Vidyavardhini's College of Engineering and Technology*
Mumbai, India
gauthamkuckian07@gmail.com

Mr. Prajjwal Vishwakarma
*Dept. of CSE(DS)*
*Vidyavardhini's College of Engineering and Technology*
Mumbai, India
prajjwalvishwakarma360@gmail.com

Prof. Krunali Vartak
*Dept. of CSE(DS)*
*Vidyavardhini's College of Engineering and Technology*
Mumbai, India
krunali.vartak@vcet.edu.in

***Abstract*-Self driving vehicles have emerged as a reliable technology that has the capability to transform transportation and mobility. The development of self-driving cars requires significant advances in a number of areas, including perception, localization, decision-making, and control. This research paper is based on the project implementation of the combination of object detection using YOLO (You Only Look Once), depth sensing using MiDaS for the localization and perception of obstacles, perspective transform and decision making for path-planning in self-driving cars. The contemporary state of the technology for object detection, depth sensing, localization and path planning evaluates the performance of the combined system through simulations and experiments. The results show that the combination of YOLO and MiDaS provides a new robust system for object detection and depth sensing. This research paper contributes to the advancement of self-driving car technology and provides new and innovative approaches to the perception and localization of obstacles in the environment.**

## I. Introduction

Self-driving cars are autonomous vehicles capable of perceiving their surrounding environment and navigating without human input. Self-driving cars have the potential to revolutionize the transportation industry, improve safety, reduce traffic congestion, and provide new mobility options for people who are unable to drive. The development of self-driving cars has been a major area of research in recent years, with the goal of creating a safer, more convenient, and more efficient transportation system. The technology behind self-driving cars involves several key components[1][2], including Sensing: uses a variety of sensors, such as radar, ultrasonic sensors, cameras, and lidars[3]. Localization: uses various methods, such as GPS, IMU, and map-based localization[3][4].Path planning: It uses algorithms to determine a safe and efficient path to the destination, taking into account the road layout, other vehicles, and other obstacles in the surrounding[5]. Control: It uses algorithms to control the vehicle's actuators, such as steering, acceleration, and brakes. There are several technical challenges associated with the development of self-driving vehicles, including robust perception and localization in complex and changing environments, safe and efficient path planning and control, and ensuring the safety and reliability of the system. Research in this area is ongoing, and significant progress is being made toward making self-driving cars a reality. Current systems can handle tasks like lane-keeping and obstacle avoidance with varying autonomy levels. Despite the hurdles, experts predict fully autonomous vehicles will become a reality in the next decade. The purpose of this research is to study the combination of object detection using YOLO, depth sensing using MiDaS for localization perspective transformation, and path planning in self-driving cars. The scope of the research includes the full architecture of self-driving cars in modules of perception, localization, and path planning, as well as an analysis of this architecture. The research involves combining YOLO and MiDaS to create a more robust and accurate system for object detection and depth sensing and evaluating the performance of the combined system through simulations and experiments. The results of the research are compared with the state of the art systems to find the potential impact and contributions of the proposed approach. In addition, the research also examines the potential challenges and limitations of the proposed system and explores avenues for future work in this area. The ultimate goal of the research is to contribute to the advancement of self-driving car technology by exploring new and innovative approaches to fully autonomous vehicles.

## II. Literature Review

### *A. Overview Of Existing Research*

In [6], an improved version of the popular YOLO object detection system. The YOLOv4 system achieves great accuracy on many benchmark datasets while conserving real time performance on modern GPUs. The authors employ several techniques to increase the accuracy percentage of object detection, including the use of larger input sizes, enhanced data augmentation, and a focus on reducing false positives. Additionally, the YOLOv4 system incorporates a new, highly efficient backbone network architecture called CSPDarknet that enables faster training and inference times.

The paper evaluates the performance of YOLOv4 on several benchmark datasets and shows significant improvements in accuracy compared to previous versions of YOLO as well as other state-of-the-art object detection systems.

The research in [7] proposed a real-time object detection algorithm called YOLO. The authors aimed to develop an algorithm that detects multiple objects in an image with good accuracy and in real-time. YOLO uses one convolutional neural network to predict bounding boxes and accurate object class probabilities directly from full images, enabling real-time object detection at 45 frames per second. In [7], authors have discussed the performance of YOLO on the PASCAL-VOC (2007 and 2012) datasets and compared it to other object detection methods. The results showed that YOLO achieved high accuracy with an average precision of 63.4% on VOC 2012 and outperformed other methods in terms of both speed and accuracy.

In [8], a novel method for real-time single image depth perception using handheld devices. The authors reviewed existing literature on depth estimation and found that most methods relied on stereo cameras or structured light sensors, which are not suitable for handheld devices. To address this limitation, the authors proposed a method that combines the machine learning techniques with the depth cues present in a single RGB image to estimate depth. The paper results also concern depth-aware AR systems, real-time and image blurring with smartphones.

The research in [9] proposes a method to calculate the intrinsic parameters of a camera by observing the projection of a quadrilateral from a single view. The need to estimate the camera parameters for 3D scene reconstruction from images motivates the paper. The proposed method utilizes the perspective projection equation and the fact that the projection of a quadrilateral retains its planarity in the image plane. By using a quadrilateral of known size and shape in the scene, the camera's focal length and the principal point can be estimated.

In [10], a multi-class 3D object recognition approach based on modern YOLO algorithms. The proposed approach aimed to improve the accuracy percentage of 3D object recognition in complex scenes by utilizing both RGB and depth information. The authors introduced a new dataset, called the NYUv2-3D, for training and testing the proposed approach. The results of the experiments showed that the proposed approach surpassed the state-of-the-art methods on the NYUv2-3D dataset, achieving mean precision of 82.8% and mean recall of 73.5%. The proposed approach in [10] also demonstrated its robustness against occlusion, noise, and cluttered background. The study highlights the potential of YOLO, in improving the accuracy of 3D object recognition in complex scenes.

The research in [11] provides a comprehensive review of the latest developments in deep learning-based object detection techniques. The authors discuss the evolution of object detection algorithms over the years and analyze the performance of various leading-edge models, including Faster R-CNN, YOLO, and SSD. They highlight the importance of several key components in object detection, such as feature extraction, the region proposal, and classification. The authors also discuss some of the challenges and weaknesses of existing object detection techniques, including occlusion, scale variation, and small object detection.

### *B. Characteristics Of Advance Self Driving Cars*

1) Improved object detection: YOLOv7 has shown significant improvements in object detection precision, accuracy, speed and efficiency making it a popular preference for full self-driving cars. By using deep neural networks, YOLOv7 can detect and track objects in real-time allowing self-driving cars to react quickly to changing traffic conditions.[11]

2) Better depth estimation: MiDaS has also shown significant improvements in depth estimation and the super-resolution allowing Full Self-Driving Cars to better perceive the environment and navigate safely. By estimating depth from a single monocular image MiDaS can provide accurate and real-time depth maps making it a promising technology for full self-driving cars.

3) Enhanced safety and reliability: Full Self-Driving Cars using YOLOv7, Midas, Perspective transform [12] and path planning algorithms exhibit the likelihood to significantly enhance safety and reliability on the roads. By accurately detecting and tracking objects, estimating depth and maneuvering through obstacles by path planning algorithms these systems can help to prevent accidents and reduce the risk of human error [13][14].

4) Technical challenges remain: Despite the promising results, there are a considerable number of technical challenges that need to be addressed before full self-driving cars using YOLOv7 and MiDaS can become a reality. These challenges encompass enhancing the accuracy and robustness of the models, speed of the whole system addressing data privacy concerns and developing better testing and validation frameworks.

5) Ethical and societal considerations: As Full Self-Driving Cars become more prevalent, there are important ethical and societal implications that must be taken into account. These include questions around legal accountability in case of collision, the potential impact on employment and urban planning, and the need for laws and norms to ensure the safe and responsible deployment of these systems.

### *C. Analysis And Limitations Of the Existing System*

The existing technology landscape for autonomous vehicles using deep learning-based object detection is very promising as demonstrated by the papers on YOLOv3, lane detection and visual odometry [15]. These approaches have shown high accuracy and faster processing times making them suitable for real-time applications in autonomous vehicles. However, certain limitations remain, including the necessity for significant amounts of labeled data to train the deep learning models and the difficulty of detecting objects in challenging lighting and weather conditions.[16] Additionally use of these deep learning models requires significant computational resources which may limit their feasibility in low-power embedded systems. The models also require continuous updates and improvements to keep up with changing environments and new objects. Further integration of these models into a complete autonomous driving system is a complex challenge that requires not only the detection of objects, localization and path planning but also control of the vehicle. Incorporating these sophisticated models into a comprehensive autonomous driving system presents a complex challenge. It requires not only object detection, localization, and path planning but also precise vehicle control, further intensifying the scope of development and integration efforts. Overall, while deep learning-based object detection, depth perception, localization via perspective transform and A-star path planning algorithm has shown significant progress and potential for self-driving cars, there are still technological, practical, and regulatory hurdles that need to be surpassed before fully autonomous driving can be achieved.

## III. Methodology

Designing a self-driving car requires a comprehensive research effort that spans multiple disciplines. This project is exactly divided into three sections and each section is implemented one after another starting with perception, plotting a occupancy grid, planning the optimal path, including various steps like data collection, training machine learning models, localization and path planning.

### A. Data Collection

Data collection is a crucial aspect of designing a self-driving car. To collect the data necessary for a self-driving car, we have two machine learning models, and each model requires its unique datas et. For object detection we have used COCO (Common Object in Context) dataset which contains 80 classes containing common objects like car, bicycle, pedestrian, etc. and divided into train and test split.[17]For depth estimation we have used MiDaS which is trained on 12 independent dataset including ApolloScape,TartanAir, IRS, KITTI, MegaDepth, ReDWeb, Movies, WSVD, HRWSI, BlendedMVS, NYU Depth V2, DIML with multi objective optimization.

### B. Training Machine Learning Models

The object detection model uses the YOLOv7 algorithm and is trained using transfer learning. The pretrained weights are loaded first and trained on COCO dataset for 100 epochs and 16 batch size. The IOU threshold is set to 30% for Non max suppression and confidence threshold is set to 25%. The best weights are saved after each epoch thus guaranteeing no further optimization. It had a mean precision of 89% and recall of 92%. Pretrained MiDaS weights are used namely dpt swin2 tiny 256 and dpt swin2 large 384 for tradeoff between speed and accuracy. With zero shot error of 8.6 on testing dataset.

### C. Localization

Object detection gives the localization in image space. The problem is that vehicles cannot directly drive onto the image space prediction. Somehow it has to be transformed to object space with real world distance. The knowledge of our vehicle with respect to the surrounding was calculated in localization using the image processing technique of perspective transform. The front view of the car was converted to a top view also referred to as the bird's eye view thus generating a 2d occupancy tracker map of 500x500 grid. Passing front view coordinates to matrices obtained by perspective transform thus creates an occupancy map, which allows obtaining the exact distance of objects. [18]

### D. Path Planning

The location of the car is obtained with respect to other obstacles in the localization module as it uses an occupancy map which keeps track of real-world distance of the ego vehicle and various obstacles present in the surrounding area. The performance of the A* algorithm was comparatively better than the other algorithms. Hence, it was the first preference. The goal point is set in top view visualization and informed search algorithm A* is used with minimum distance heuristic to predict and traverse through obstacles to the goal point.[19]

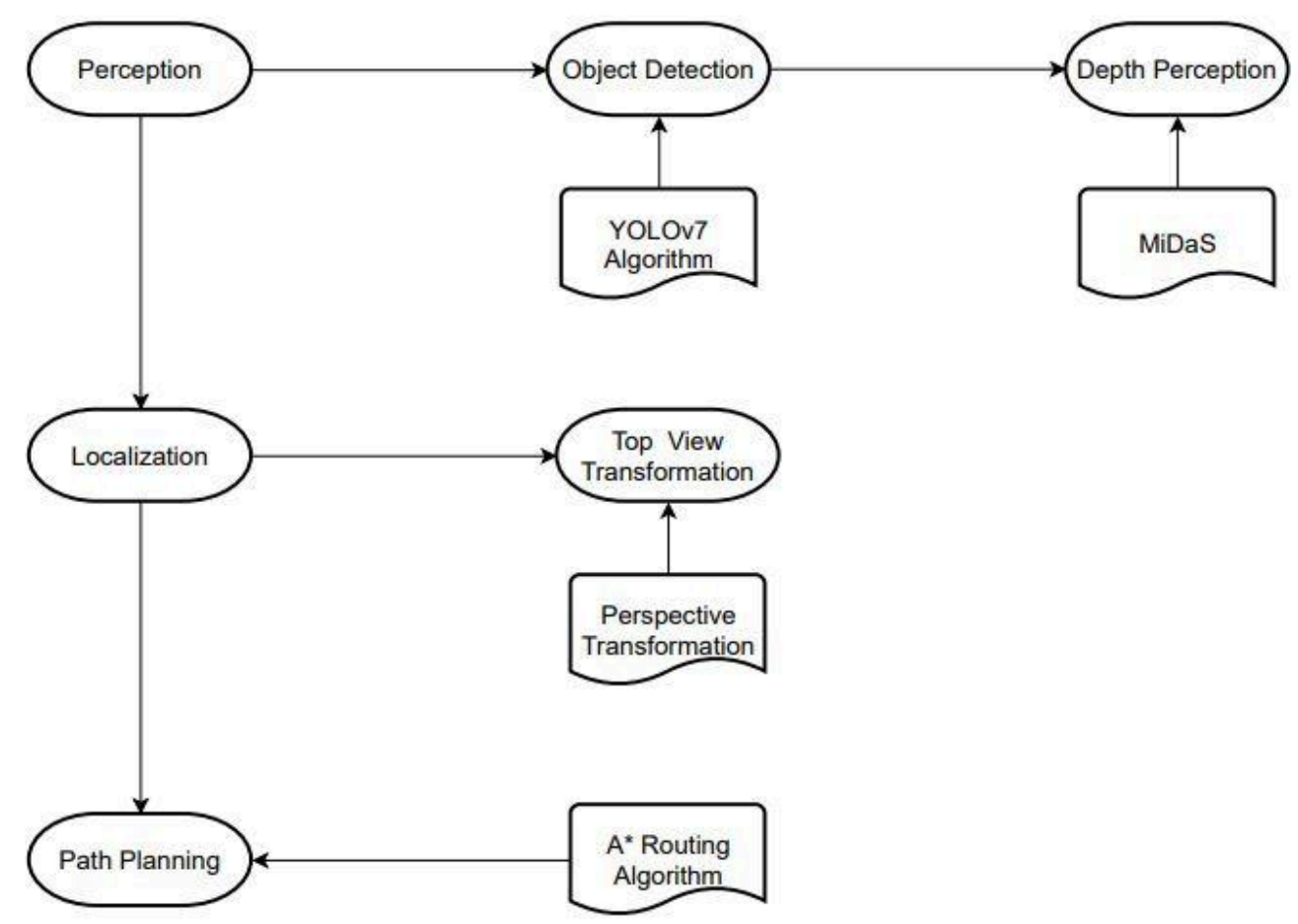

Fig. 1. Architecture of system

## IV. Results

The objective of this analysis is to evaluate the performance of self-driving vehicles in the whole system. The available dataset used in this analysis was collected during a series of tests in various driving conditions, including different lighting and weather.[20]

### A. Object Detection

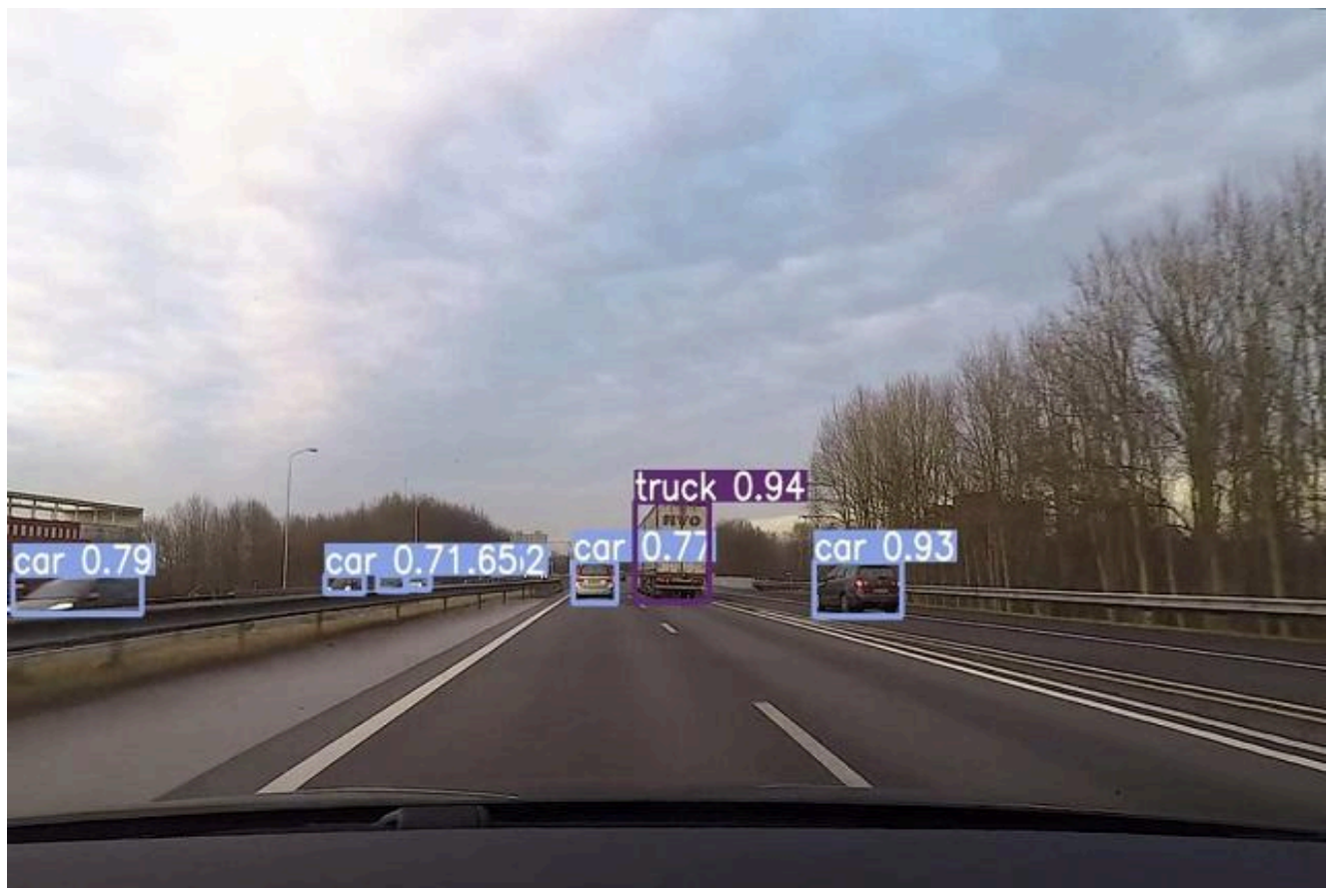

Fig.2. YOLO Object Detection

Our analysis shows that the object detection system had a mean precision of 89% and a mean recall of 92%. This means that the system was capable of accurately detecting and classifying objects most of the time. The system performed better in well-lit and clear weather conditions, with a precision of 93% and a recall of 95%.However, in low-light and rainy conditions, the system's performance dropped to a precision of 83% and a recall of 87% with the frame rate of 80 frames per second on GeForce RTX3060.

### B. Depth Sensing

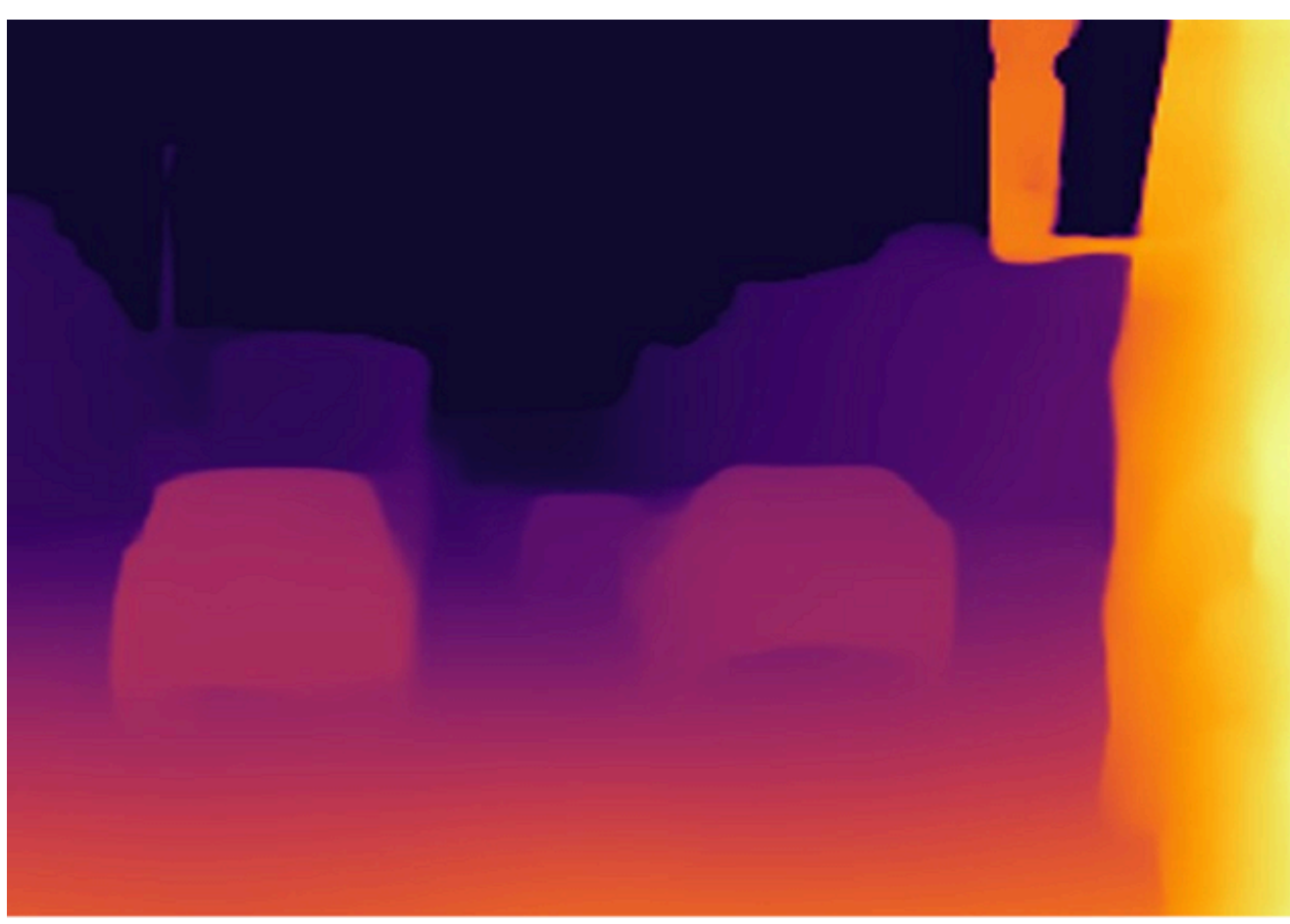

Fig. 3. MiDas Depth Perception

Our analysis also evaluated the object depth tracking system's performance. The depth sensing system was able to run in real-time with an average frame rate of 74 FPS on the same system. The system's performance was slightly varied across different lighting and weather conditions with an average 8.6 zero shot error[21].

### C. Localization

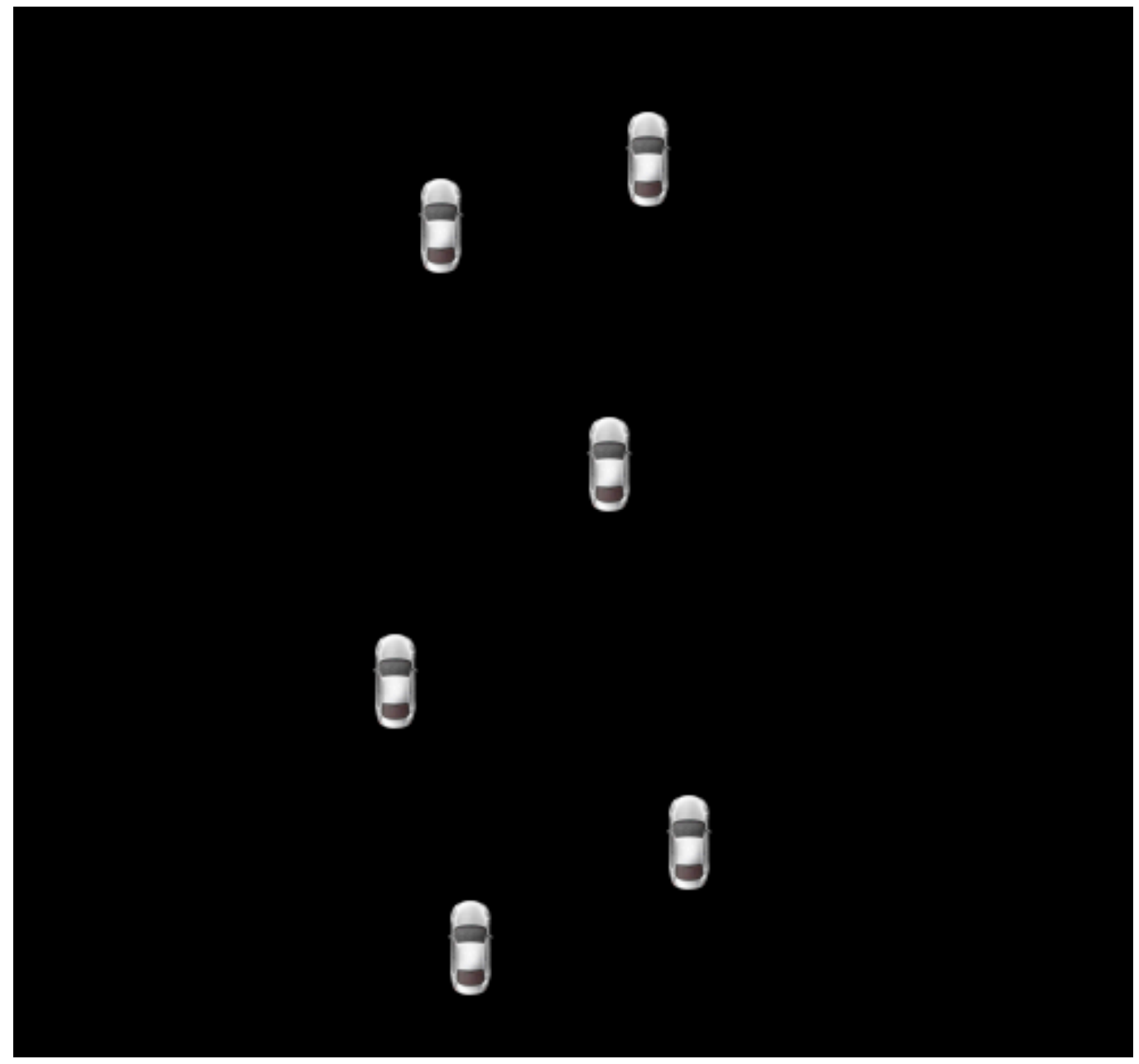

Fig. 4. Top View Localization

The 2D occupancy tracker map locates our vehicle with respect to the surrounding by making a 2D visualization of it as shown in the front facing view and can rapidly plot a continuously changing surrounding. The size of the grid 500x500 was found to be the best suited for the tradeoff between accuracy and latency on our system. The grid size smaller than this resulted in a poor path quality due to less nodes to explore while grid size bigger than this increased computational power thus not suited for real time.

### D. Path Planning

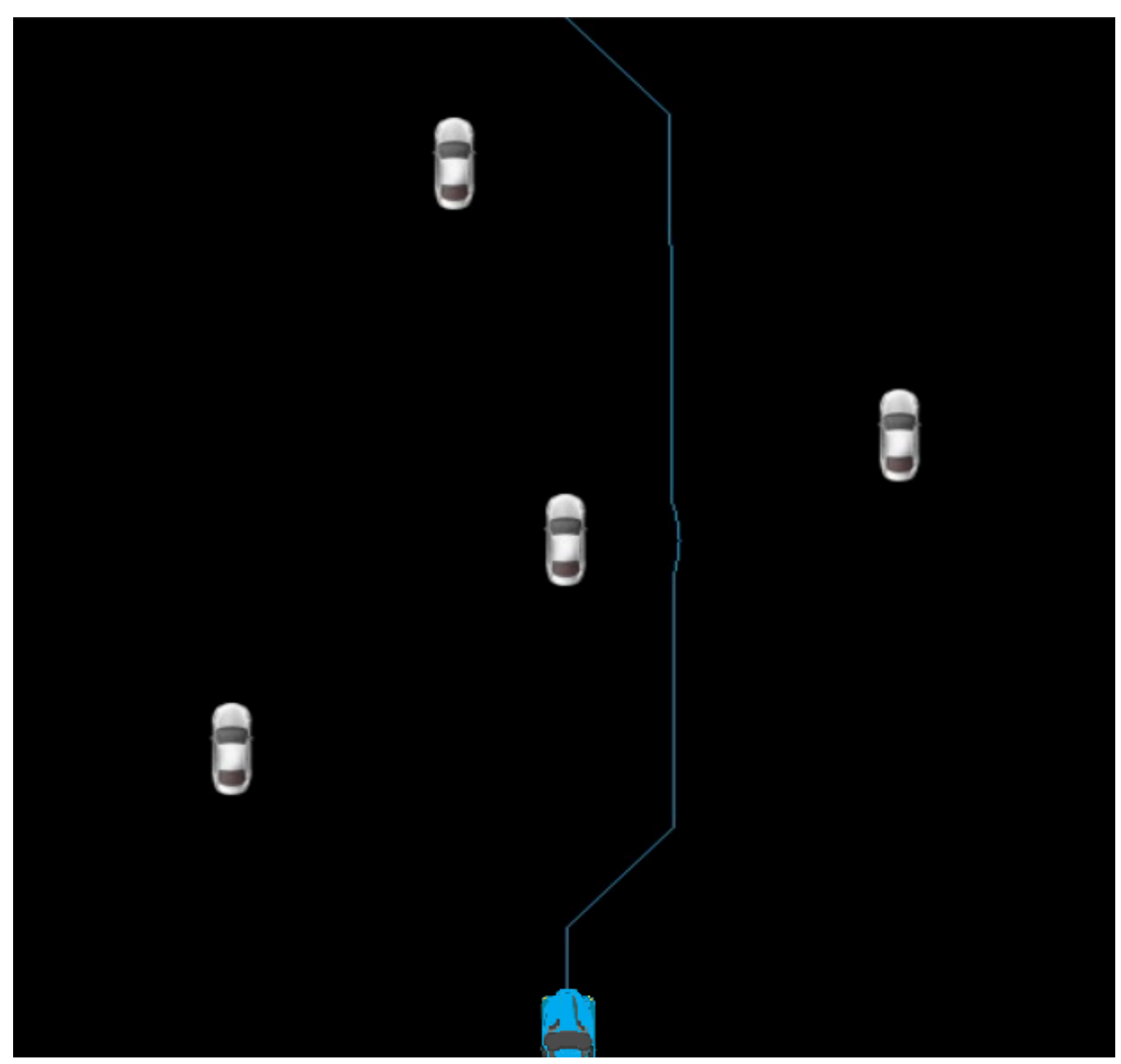

Fig. 5. Path Planning

In order to compare the performance and effectiveness of different path planning algorithms, a comparative analysis was conducted using test cases of various surroundings. The algorithms under consideration included A* (Diagonal distance heuristic), BFS, Dijkstra, and DFS. Quantitative measures including path length, execution time, number of expanded nodes, and max turning angle, were used to measure the performance and capabilities of these algorithms. Multiple test cases were designed to represent different simulation scenarios, including environments with varying obstacle densities. Each algorithm was executed on these test cases on our system, and the results were tabulated as shown below.

TABLE I. ALGORITHM PERFORMANCE COMPARISON

| Indicators | A* | BFS | Dijkstra | DFS |
|---|---|---|---|---|
| Path Length (pixels) | 504 | 504 | 504 | 1865 |
| Execution Time/s | 0.9281 | 0.9091 | 0.9295 | 0.7436 |
| Number of Expanded Nodes | 49968 | 124920 | 134914 | 1865 |
| Max Turning Angle(°) | 45° | 135° | 45° | 90° |

## V. DISCUSSION

### A. Limitations And Challenges Of the Research

Some potential limitations and challenges that may be associated with research:

1) Dependence on object detection: The research project has a heavy dependence on object detection. In object detection, the main aim is to detect objects and algorithms that can only detect objects on which they are trained. As our model is trained on a COCO dataset having 80 classes our model can only detect 80 different objects. Any object which is not included in the dataset used will fail to detect and result in collision [22].

2) Monocular Depth estimation: The depth estimation using a monocular camera is almost impossible. With a single monocular camera, it is challenging to accurately deduce the real depth of objects in the scene without an additional depth cue. MiDaS often rely on features such as edges, corners, or texture patterns to infer depth information. It is also prone to scale ambiguity. Without an absolute scale reference, difficulty arises in determining the true size or absolute distance of objects accurately.

3) Single Front facing camera: The limitation of using a single front-facing camera is that it has a limited field of vision, making it difficult to detect objects and road conditions outside its range. It cannot see the whole 360° around the car making it more vital from both sides and back of the car [23]. Dependence of the whole system on just one camera can be dangerous. If the camera malfunctions or encounters any technical issues, it can significantly impact the perception capabilities of the self-driving car, leading to potential safety concerns.

4) Restricted to local path planning: As this research does not have access to a global map of the environment it is just limited to local path planning. The goal point for this research is hardcoded and assumed at the horizon in front of the car. For real world applications there should also be integration of global path plans [24].

5) Limited testing conditions: Systems prototypes have been tested in controlled conditions which may not reflect the complexity and unpredictability of real-world driving scenarios [25]. There are no mechanisms to test these systems on diverse and unpredictable scenarios which often result in failure in the real world. Simulations designed for specific tasks fail to provide unpredictability conditions which autonomous vehicles may face after deployment, making it not a fully reliable source of testing the environment.

6) Safety concerns: Safety is a critical consideration for this system and any issues or failures could have serious consequences. Engineers and researchers continuously subject systems to various simulated scenarios and controlled environments to assess their responses and identify potential weak points. Also testing public roads presents a significant risk and requires extensive safety protocols and regulations. The pursuit of a safe navigation mechanism is an ongoing journey in the realm of self-driving cars.

7) Ethical consideration: Range of ethical considerations are raised such as how to prioritize safety in different scenarios and how to allocate responsibility in the event of accidents or failures. Additionally, questions about the decision-making process of autonomous cars concerning harm to occupants versus pedestrians and the handling of moral dilemmas come to the prime subjects of the ethical discussion.

### B. Suggestions For Future Research

Some potential suggestions for future research in the field of autonomous cars:

1) Robust techniques for localization: Other localization techniques like SLAM and Visual Odometry can be integrated with current systems for robust localization. By fusing multiple sensor sources, self-driving cars can achieve more accurate and resilient location, enabling them to move through complex environments and difficult scenarios with greater safety and knowledge.

2) Faster overall architecture: The main objective of future research must be to develop fast architecture of the overall project as time is a crucial component for the problem of autonomous driving. By creating low latent and streamlined systems, researchers can address the difficulties of processing vast amounts of input sensor data, machine learning models, complex algorithms, and decision making in a time sensitive manner, thus propelling the field of autonomous driving to a great extent and utilizing its full potential.

3) Better handling of adverse weather conditions: Adverse weather conditions such as fog, snow and rain can pose challenges for self-driving cars. Future research could focus on developing algorithms and sensor systems that can handle these conditions more effectively. There should not be any reliability of performance on a system driven by outside weather conditions, these systems should navigate through the environment irrespective of these adversarial conditions.

4) Integration of self-driving cars with an existing transportation infrastructure: Self-driving cars are integrated into an existing transportation infrastructure, including roadways and traffic systems. Future research should focus on developing in depth strategies that enable self-driving vehicles to co-exist harmoniously with traditional vehicles, pedestrians, and cyclists on roadways. This includes the development of communication protocols to facilitate smooth interactions with traffic signals, smart intersections, and other traffic management systems Future research could focus on developing strategies for effectively integrating self-driving cars into this infrastructure.

## VI. Conclusion

Through the innovative fusion of YOLO, MiDaS, localization using perspective transformation, and the A* a routing algorithm for path planning, our approach introduces a distinct solution that demonstrates alternative approaches to common problems in the field of self-driving systems. Through extensive experimentation and analysis, our object detection model has shown promising results with average precision of 89% and an average recall of 92%. This means that the system was able to accurately detect and classify objects in most cases. Our distance estimation has an average zero shot error of 8.6 is slightly better than the error rates reported in previous versions of MiDaS. However, it is important to note that despite the improvements achieved, our model is still susceptible to errors caused by various factors such as adverse weather conditions and limitations in camera quality and also possess some of the limitations as discussed. While acknowledging the existing body of research on similar models, our study builds upon these foundations by introducing comprehensive analysis and variations. With further dedicated research and refinement, our approach offers a valuable benchmark for future development of autonomous vehicles. Future research directions could focus on further optimizing the model's performance, exploring additional techniques for feature-extraction or incorporating newer technologies to enhance its capabilities.